\documentclass[sn-mathphys]{sn-jnl}% Math and Physical Sciences Reference Style
\usepackage{tabularx,booktabs}
\usepackage{url}
\usepackage[figuresright]{rotating}
\usepackage{diagbox}
\usepackage{adjustbox}
\newcolumntype{C}{>{\centering\arraybackslash}X}
\jyear{2021}%

\theoremstyle{thmstyleone}%
\theoremstyle{thmstyletwo}%

\theoremstyle{thmstylethree}%

\begin{document}

%\title[Article Title]{A Survey on Computer Vision Based Behavior Assessment to Recognize Animal Pain}

\title[Going Deeper than Tracking]{Going Deeper than Tracking: a Survey of
Computer-Vision Based Recognition of Animal Pain and Affective States}


\author*[1]{\fnm{Sofia} \sur{ Broom\'e}}\email{sbroome@kth.se}
\equalcont{These authors contributed equally to this work.}
\author[2]{\fnm{Marcelo} \sur{Feighelstein}}\email{feighels@gmail.com}
\equalcont{These authors contributed equally to this work.}

\author[2]{\fnm{Anna} \sur{Zamansky}}\email{annazam@is.haifa.ac.il}
\equalcont{These authors contributed equally to this work.}

\author[3]{\fnm{Gabriel} \sur{Carreira Lencioni}}\email{gabriel.lencioni@gmail.com}
\equalcont{These authors contributed equally to this work.}

\author[6]{\fnm{Pia} \sur{Haubro Andersen}}\email{pia.haubro.andersen@slu.se}

\author[7]{\fnm{Francisca} \sur{Pessanha}}\email{f.pessanha@uu.nl}

\author[4]{\fnm{Marwa} \sur{Mahmoud}}\email{marwa.mahmoud@glasgow.ac.uk}

\author[1,5]{\fnm{Hedvig} \sur{Kjellstr\"om}}\email{hedvig@kth.se}

\author[7,8]{\fnm{Albert Ali} \sur{Salah}}\email{a.a.salah@uu.nl}

\affil[1]{\orgdiv{Division of Robotics, Perception and Learning}, \orgname{KTH Royal Institute of Technology}, \orgaddress{ \city{Stockholm}, \country{Sweden}}}

\affil[2]{\orgdiv{Tech4Animals Lab, Information Systems Department}, \orgname{University of Haifa},\orgaddress{ \city{Haifa}, \country{Israel}}}

\affil[3]{\orgdiv{Department of Preventive Veterinary Medicine and Animal Health}, \orgname{School of Veterinary Medicine and Animal Science, University of São Paulo}, \orgaddress{ \city{São Paulo}, \state{SP}, \country{Brazil}}}

\affil[4]{\orgdiv{School of Computing Science}, \orgname{University of Glasgow}, \orgaddress{ \city{Glasgow}, \country{United Kingdom}}}

\affil[5]{\orgname{Silo AI}, \orgaddress{ \city{Stockholm}, \country{Sweden}}}

\affil[6]{\orgdiv{Department of Clinical Sciences}, \orgname{Swedish University of Agricultural Sciences}, \orgaddress{ \city{Uppsala}, \country{Sweden}}}

\affil[7]{\orgdiv{Department of Information and Computing Sciences}, \orgname{Universiteit Utrecht}, \orgaddress{\city{Utrecht}, \country{Netherlands}}}
\affil[8]{\orgname{Boğaziçi University}, \orgaddress{ \city{Istanbul}, \country{Turkey}}}

%%
\abstract{
Advances in animal motion tracking and pose recognition have been a game changer in the study of animal behavior. Recently, an increasing number of works go `deeper' than tracking, and address automated recognition of animals' internal states such as emotions and pain with the aim of improving animal welfare, making this a timely moment for a systematization of the field. This paper provides a comprehensive survey of computer vision-based research on recognition of affective states and pain in animals, addressing both facial and bodily behavior analysis.  We summarize the efforts that have been presented so far within this topic -- classifying them across different dimensions, highlight challenges and research gaps, and provide best practice recommendations for advancing the field, and some future directions for research.}

\keywords{Affective computing, non-human behavior analysis, pain estimation, pain recognition, emotion recognition, computer vision for animals}


\maketitle

\section{Introduction}

It is now widely accepted that animals can not only suffer and feel pain \cite{Sneddon2014}, but also experience positive emotional states~\cite{de2016we, birch2021review}. Although traditionally, animal welfare science focus has been on pain and suffering, a recent paradigm shift is also addressing quality of life in a broader sense, seeking an understanding of animals' positive affective experiences \cite{duncan1996animal,boissy2007emotions}. %Therefore, methods that can aid the assessment of animals' affective experiences are needed.

There is no common agreement on what constitutes animal emotions (see~\cite{paul2018animal,kret2022my} for comprehensive reviews). However, emotions are often described as internal states which are expressed in physiological, cognitive and behavioral changes~\cite{anderson2014framework}.
Pain, traditionally studied separately from emotions, also has an affective component and is described as ``an unpleasant sensory and emotional experience” \cite{raja2020revised}.

Due to the subjective nature of affective states, their identification and measurement is particularly challenging, especially seen
the lack of verbal basis for communication in non-human animals. To address this problem, physiological, cognitive, or behavioral changes that occur whenever animals experience affective states are often 
used as putative indicators \cite{mendl2010integrative}.

While physiological and cognitive changes are difficult to
observe, measuring behavior is considered one of the most promising and {least} invasive methods for studying affective states.  
 It is widely agreed that facial and body expressions may convey information on emotional states \cite{descovich2017facial,diogo2008fish}, including the affective components of pain. {These expressions are produced and used for communication by most mammalian species \cite{diogo2008fish,briefer2015emotions, schnaider2022vocalization, seneque2019could}.}

Traditional methods for measuring behavior are based on either direct observation or by video recording analysis of one or more subjects, documenting carefully pre-designed behavioral categories, designed to be as unambiguous as possible \cite{bateson2021measuring}. The categories and ethograms may be designed for specific purposes. 
%The Facial Action Coding System (FACS)~\cite{Ekman1978}, is an ethogram developed for measuring changes in the facial expressions, and the Body Action and Posture Coding System (BAP)~\cite{dael2012body} is developed for coding body movements in behavioral research. 
 
For objective measurement of facial expressions of humans, the Facial Action Coding System (FACS) was developed to describe movements of facial muscles in terms of facial action units (AUs)~\cite{Ekman1978FacialAC}. Likewise, the Body Action and Posture Coding System (BAP)~\cite{dael2012body} was designed for coding body movements in human behavioral research. 
With the same goal, %In a similar matter, 
coding systems were developed for other animals for both face~\cite{correia2021extending,waller2013dogfacs,wathan2015equifacs,caeiro2017development} and body%~\cite{}
.

% the gold standard for measuring ‘action units’, assigning codes to activity of individual muscles or groups of muscles. It has recently been adapted for several non-human species, including several non-human primates~\cite{correia2021extending}, dogs~\cite{waller2013dogfacs},  horses~\cite{wathan2015equifacs}  and cats~\cite{caeiro2017development}. A simplified method, applied mainly in the context of pain, are grimace scales \cite{mogil2020development}. They are species-specific and have been developed for several species, such as mice~\cite{langford2010coding}, rats~\cite{sotocina2011rat}, cows~\cite{gleerup2015pain}, lambs~\cite{guesgen2016coding} horses~\cite{dalla2014development}and cats~\cite{evangelista2019facial}. In addition to facial expressions, other  behavioral indicators have also been studied in order to assess pain, discomfort and affective states in animals, while also differentiating between the valence of each affective states, from negative to positive{~\cite{ede2019symposium, hall2018assessing, lansade2018facial}.}

Methods based on human observation and manual coding carry several serious limitations. They often require extensive human training, %for the specific ethogram protocol in question
as well as rater agreement studies for reliable application. Furthermore, they are time consuming, and
%observation times are short and 
prone to human error or bias \cite{anderson2014toward}. Computational tools, and especially tools based on computer vision, provide an attractive alternative \cite{anderson2014toward}, since they are non-invasive, enable 24 hours a day surveillance, save human effort and have the potential to be more objective than human assessments \cite{andersen2021towards}.
%The human observational research is therefore no longer a sufficient approach for the
%development of methods that can ascertain the emotional welfare aspects of animal
%lives. 

%Automation of behavior measurement, and specifically computer vision based approaches can provide an attractive alternative.

%\GL{ Although several methods have been developed recently to assess pain or other indicators related to affective states in animals, all of those methods require adequate training and can be time consuming, which makes this practice underused on daily routine at animal hospitals, farms, equestrian centers, etc. Computer vision systems might be a promising approach since it would be available to monitor animals full time and without relying on the availability of a trained professional. The most important aspect of those systems is that they could possibly contribute to a new understanding of the studied behaviors, therefore gathering relevant data which are usually lost by human observers.}

%In the human domain, automated facial and body gesture analysis is a rapidly expanding field of research.

In the human domain, automated facial and body gesture analysis is a rapidly expanding field of research. Accordingly, many datasets are available with extensive annotations for emotional states are available.
%, with self-assessment questionnaires being a common form of annotation.
%This is largely owing to %this owes to 
%computer vision-based approaches, which have
%access to large annotated datasets, %due the non-invasive nature,
%where
%since the possibility to
%verbally communicate the emotional state facilitates annotation. 
%which has produced a huge body of work pushing forward both emotion and pain research. Among the various developed approaches, %computer-vision based ones are among the most popular due to their non-invasive nature.{computer vision-based ones are popular owing to their non-invasive nature.} 
Comprehensive surveys cover analysis of facial expressions~\cite{li2020deep}, and body gestures~\cite{noroozi2018survey}, with the recent trend being to combine several modalities using multi-modal emotion recognition approaches~\cite{sharma2021survey}. In the context of pain, numerous works have addressed facial expression assessment in humans \cite{MAlEidan2020DeepLearningBasedMF, Hassan2021AutomaticDO}, and, notably, in infants \cite{zamzmi2017review}. %; the review by Zamzmi et al.~\cite{zamzmi2017review} identifies facial expressions as one of the most common pain indicators in infants.

Although %animal-related fields
research concerned with non-human animal behavior have so far lagged behind the human domain with respect to automation, recently, the field is beginning to catch up. This is owing in part to developments in animal motion tracking with the introduction of general platforms, such as DeepLabCut \cite{mathis2018deeplabcut}, EZtrack \cite{pennington2019eztrack}, Blyzer\cite{amir2017k9}, LEAP \cite{pereira2019fast}, DeepPoseKit \cite{graving2019deepposekit} and idtracker.ai \cite{romero2019idtracker}. However, as pointed out in \cite{forkosh2021animal}, ``being able to track the intricate movements of a spider, a cat, or any other animal does not mean we understand its behavior". %As well as
Similarly, presenting good rater agreement
on a given behavior does not mean that the behavior actually measures %measures
a given emotion. Automated recognition of affective states and pain is an important and difficult problem that requires going deeper than tracking movements, to assess whether the observable behaviors in fact correspond to internal states. %, addressing the subsequent analysis of facial expressions and body language that serve as indicators for internal states.
Such analysis of facial expressions and body language brings about challenges that are not present in the human domain, as described in \cite{hummel2020automatic}. In particular, these are related to data collection, ground truth establishment and a great variety of morphological differences, shapes and colors within and across animal species. 

Indeed, in recent years the number of vision-based research articles addressing these topics is growing. To promote systematization of the field, as well as to provide an overview of the methods that can be used as a baseline for future work, this paper aims to provide a comprehensive survey of articles in this area, focusing on the visual modality. 
Based on the survey, we also provide technical best-practice recommendations for future work in the area, and propose future steps to further promote the field of animal affective computing. 

\section{Survey Scope and Structure}
This survey covers works addressing automated recognition of affective states and pain in animals using computer vision techniques. This means that many interesting works related to automation of recognition of behavior in animals are left out of scope, including the large and important fields of animal motion tracking, precision livestock farming, methods for landmarks detection and 3D modeling of animal shapes. Moreover, due to our vision focus, we only consider research focused on analysis of images or video, excluding works that are based on audio, wearable sensor data, or physiological signals.

%\subsection{Survey Structure}

We begin with an overview of relevant background within affective states and pain research in non-human mammals in Section~\ref{sec:veterinary_bg}. This section will be rather condensed, as previously published work in biology covers this topic well. We provide pointers to this literature and summarize the findings. 
% (Section \ref{section:main}). In doing so, we identify three axes that are especially relevant for the application at hand: 1) using information from facial expressions or whole body pose and movements, 2) the amount of temporal information that a method uses, and 3) the amount of pre-defined cues and hand-crafted engineering that is done before high-level analysis such as, e.g., pain classification. The review will be oriented around these three aspects. Finally, we discuss the further challenges and open problems.
Section~\ref{sec:review} provides a meta-analysis of 
% major works falling under the scope described above, i.e., presenting a 
computer vision-based approaches for classification of internal affective states in non-human animals. We include articles that perform facial action unit recognition, since this task is closely connected to pain or affective states assessment. The articles included were identified by web search using the terms `automated animal pain/affect/emotion recognition/detection', `pain/affect/emotion recognition/detection in animals', `computer vision based recognition of pain/affect/emotion in animals', and by tracing references of the different works.

We dissect these works according to the different workflow stages: data acquisition, annotation, analysis, and performance evaluation, respectively. For each of these steps, we identify and highlight the different approaches taken together with common themes and challenges. Based on our dissection and drawing parallels with human affective computing research, Section~\ref{sec:best} provides some best practice guidance for future work related to technical issues, such as data imbalance, cross-validation and cross-domain transfer. 
Section~\ref{sec:discussion} draws further conclusions from our analysis, 
identifying crucial issues that need to be addressed for pushing the field forward, and reflects on future research directions. 

%go on to present the content of the research on computer vision based pain and affective states recognition works that have been presented so far. Table \ref{table:works} presents the works {included} 
%on which we focus 
%in this survey, {together with} their classification along the dimensions of species, internal state recognized, and type of behavioral indicators measured. {In what follows, we }%We then
%proceed to analyze the {articles} %works
%according to {a standard } %the typical
%process 
%of addressing the problem at hand: data acquisition and annotation (Section \ref{sec:data}), data analysis, {including} %where 
 %models, feature selection {where applicable}, and overall pipelines (Section \ref{sec:analysis}), and performance evaluation (Section \ref{sec:evaluation}). For each of these stages, we identify important subdimensions across which we further compare the works.  

% How it is done
\section{Research on Affective States and Pain in Non-human Animals}
\label{sec:veterinary_bg}
% for example body/face,  affective state/pain, grimace scale/facs
%\subsection{Animal emotions }
% Emphasize importance of computer vision because less invasive

In 2012, the Cambridge declaration on consciousness was signed, stating that “The absence of a neocortex does not appear to preclude an organism from
experiencing affective states". This implies that in addition to all mammals, even birds and other species, such as octopuses and crustaceans potentially experience emotions~\cite{birch2021review, low2012cambridge}. Affective states incorporate 
emotions, mood %, feelings
and other sensations, including pain, that have the property of valence~\cite{mendl2020animal}. Although the expression of emotions has been heavily discussed, there is no clear-cut definition of each of those terms,  
especially when referring to non-verbal individuals. A common approach is to consider them as intense, short-term affective states triggered by events in which reinforcers (positive reinforcers or rewards, and negative reinforcers or punishers) are present or expected~\cite{paul2018animal,dawkins2008science}.

There are also different approaches for the classification of emotional states, with one of the most prominent being the discrete one. According to this theory, animals have a certain number of fundamental emotion systems, based on neuronal structures of different brain areas homologous across species~\cite{panksepp2010emotional}, leading to a discrete set of distinct emotional states. Paul Ekman, for instance, described the following six distinct emotions in humans: fear, sadness, disgust, anger, happiness, surprise~\cite{ekman1992argument}, but other discrete classifications have been suggested as well (e.g., by Panksepp~\cite{panksepp2010emotional}). Plutchik~\cite{plutchik1980emotion} further suggested that emotions should provide a function aiding survival, which is relevant, but difficult to use as a working definition. An alternative classification system is the dimensional approach, where the affective
states are classified according to their valence (negative or to positive -- bad
or good) and arousal levels~\cite{mendl2020animal, posner2005circumplex}. Anecdotal evidence that non-human
animals can experience secondary emotions as grief, jealousy and more, is
compelling~\cite{morris2008secondary, uccheddu2022domestic}. According to this view, several affective states could be manifested at the same time, making the recognition of affective states in non-human animals even more challenging.

%to ascribe certain behaviors or behavioral patterns to a specific affective
%state, a major research challenge in non-verbal individuals is consider if other
%affective states are present at the same time.

Pain research developed separately from emotion research in both human and non-human animals, despite the close links between them \cite{hale1997emotional}. According to the International Association for the Study of Pain, human pain is defined as ``an unpleasant sensory and emotional experience associated with actual or potential damage, or described in terms of such damage"~\cite{raja2020revised}; thus, the emotional dimension of pain can be considered an affective state. 

Pain assessment in infants is considered one of the most challenging problems in human pain research \cite{anand2007pain}, due to the issue of non-verbality in neonates and older infants, analogously to non-human animals. Historically, even the ability of human neonates to feel pain and emotions was questioned \cite{grunau1987pain, camras2010emotional}. As late as the 1980s, it was widely assumed that neonates did not experience pain as we do, together with a hesitancy to administrate opiates to these patients, and surgery was often performed without anaesthesia \cite{Fitzgerald1989PainAA}. Duhn and Medves \cite{duhn2004systematic} provide a systematic review of instruments for pain assessment in infants, where facial expressions are identified as one of the most common and specific indicators of pain. Facial
expression of pain in neonates is defined as the movements and distortions in facial muscles associated with a painful stimulus; the facial
movements associated with pain in infants include deepening of the nasolabial furrow, brow lowering, narrowed eyes, chin
quiver and more \cite{zamzmi2017review}.

%The valence of concurrent emotional states and external input can therefore influence how the animal perceives painful stimuli \cite{loggia2008experimentally}. 

The assessment of pain and affective states in non-human mammals is much less explored than in the human domain, due to the difficulties regarding ground truth and subsequent lack of large databases mentioned above. The pressing need for these assessments, be it pain, stress or positive emotions in animal health and welfare evaluations, has therefore made researchers resort to addressing measurements 
%The assessment of pain and affective states in non-human animals is much less explored than in the human domain.
%As with the case of neonates, it is also  extremely challenging, as the lack of verbal communication rules out commonly methods based on self-reporting in human emotion \cite{barrett2004feelings} and pain research \cite{labus2003self}. 
%Thus researchers commonly resort to addressing measurement 
of physiological, behavioral, and cognitive components of affective states, which can be measured objectively, and even in many cases automatically \cite{paul2005measuring,kret2022my}. This involves physiological (such as heart rate, hormone levels, body temperature) and behavioral parameters (vocalisations, facial expressions, body postures). Naturally, behavioral parameters are particularly relevant when exploring computer vision-based approaches. 

Facial expressions, produced by most mammalian species \cite{diogo2008fish} are one important source of information about emotional states \cite{descovich2017facial,diogo2008fish}. Behavioral parameters such as facial expressions are not only non-invasive to observe, but have also proved more reliable than physiological indicators. The latter are significantly influenced by diseases and can only be used in controlled settings \cite{andersen2021towards,gleerup2016recognition}. Adaptations of FACS to several other species have been used for measuring facial behavior in non-human primates~\cite{correia2021extending}, dogs~\cite{waller2013dogfacs},  horses~\cite{wathan2015equifacs} and cats~\cite{caeiro2017development}. Grimace scales can be less demanding to apply than FACS, since they analyze movements and behavior changes in a small set of facial regions related to pain \cite{mclennan2019development, andersen2021towards}.
Further, facial behavior such as eye blink rates and twitches \cite{merkies2019eye, mott2020blink}, as well as yawning, have been related to stress and stress handling. A correlation between positive emotions and animal facial behavior has also been shown; as an example, in cows, the visibility of the eye sclera dropped during positive emotional states \cite{proctor2015measuring}.

%Affective states have physiological, behavioral and cognitive components \cite{kret2022my}. In the animal domain, pain recognition has received special attention due to its applications for pain management and welfare assessment. 

%The Facial Action Coding System (FACS)~\cite{Ekman1978}, is the gold standard for measuring facial movements or ‘action units’, assigning codes to activity of individual muscles or groups of muscles. It has recently been adapted for several non-human species, including several non-human primates~\cite{correia2021extending}, dogs~\cite{waller2013dogfacs},  horses~\cite{wathan2015equifacs}  and cats~\cite{caeiro2017development}. A simplified method, applied mainly in the context of pain, are grimace scales. They are species-specific and have been developed for several species, such as mice~\cite{langford2010coding}, rats~\cite{sotocina2011rat}, cows~\cite{gleerup2015pain}, lambs~\cite{guesgen2016coding} horses~\cite{dalla2014development}
%and cats~\cite{evangelista2019facial}. 

In addition to facial expressions, other behavioral indicators have been studied in order to assess pain, discomfort, and affective states in animals. These parameters have also been used to estimate the valence of each affective state, from negative to positive{~\cite{ede2019symposium, hall2018assessing, lansade2018facial}. 
Similarly to facial behavior, body posture and movement have been correlated to a range of affective states and pain-related behavior~\cite{walsh2014pain, seneque2019could, dyson2018development, briefer2015emotions, schnaider2022vocalization}. 
Further, several protocols have also been developed to assess behavioral indicators such as changes in consumption behaviors (time activity budgets for eating, drinking, or sleeping, etc.) \cite{oliveira2022hospitalisation,auer2021activity, maisonpierre2019accelerometer}, anticipatory behaviors, affiliative behavior, agonistic behaviors, and displacement behaviors, amongst others.}

%visualization of the sclera (white of the eye)~\cite{proctor2015measuring}, eye wrinkles~\cite{hintze2016eyes} and blink rate~\cite{merkies2019eye, mott2020blink}, tail posture and movement, yawning \cite{miller2010handling}, chewing and vocalizations~\cite{yeon2012acoustic}. 

Other behavioral assessments, such as tests (Open field, Novel object and Elevated plus maze) \cite{lecorps2016assessment}  and Qualitative Behavior Assessment (QBA) have also been used \cite{kremer2020nuts}. Less used
is the Body and Posture coding system, which recently was adapted for use in dogs and
horses \cite{rashid2020equifacsplos,lundblad2021effect}. 
The advantage of using an ``exhaustive'' coding scheme is that the coding can be done
without any anticipation of what will be detected. In contrary, in a pain score scale, pain is
always the issue. %This means that such a scale can only determine what is believed to
%happen. 

When carefully coding facial actions of horses in pain and horses with emotional stress from isolation, it became obvious that pain is associated with some degree of emotional stress. Additionally, stress without pain may have some similarities to pain during the acute stages \cite{lundblad2021effect}. External inputs that may induce stress can therefore influence prototypical facial expressions. The mixing problem holds for other affective states as well, adding to the challenge of recognizing specific internal states. 
%same is true for other affective states that might occur simultaneously, therefore making it difficult to assess whether an animal is experiencing one or more affective states at a time. 
%Other affective states may potentially occur simultaneously, 
Because of these challenges, assessing affective states such as the ones resultant from stressful situations may not be as direct as pain recognition, for which there are several validated indicators  \cite{lundblad2021effect, mayo2019face}. This might also be the reason for the lack of automated methods regarding this matter.

\section{Meta-analysis of Computer Vision-Based Approaches for Classification of Affective States and Pain}
\label{sec:review}

\begin{sidewaystable}[]
\caption{A summary of the reviewed works, categorized by the state in question, potential stimulus, focus area, whether the work includes a state classifier and whether the state annotations are behavior- or stimulus-based.}
\label{tab:works}
\resizebox{\textwidth}{!}{
\begin{tabular}{@{}lllllccc@{}}
\toprule
\multicolumn{1}{c}{\multirow{2}{*}{\textbf{Study}}} & \multicolumn{1}{c}{\multirow{2}{*}{\textbf{Species}}} & \multicolumn{1}{c}{\multirow{2}{*}{\textbf{State}}} & \multicolumn{1}{c}{\multirow{2}{*}{\textbf{Stimulus}}} & \multicolumn{1}{c}{\multirow{2}{*}{\textbf{Focus area}}} & \multirow{2}{*}{\textbf{State classifier}} & \multicolumn{2}{c}{\textbf{State annotations}} \\ \cmidrule(l){7-8} 
\multicolumn{1}{c}{}                                & \multicolumn{1}{c}{}                                  & \multicolumn{1}{c}{}                                & \multicolumn{1}{c}{}                                   & \multicolumn{1}{c}{}                                     &                                            & \textbf{Behavior-based}  & \textbf{Stimulus-based}  \\ \midrule
Tuttle et al.~\cite{tuttle2018deep}                 & \multirow{2}{*}{Mice}                                 & \multirow{2}{*}{pain}                               & vet. procedure                                         & face                                                     & +                                          & \checkmark               & \checkmark               \\
Andresen et al.~\cite{andresen2020towards}          &                                                       &                                                     & vet. procedure                                         & face                                                     & +                                          & \checkmark               & \checkmark               \\ \midrule
Mahmoud et al.~\cite{mahmoud2018estimation}         & \multirow{2}{*}{Sheep}                                & \multirow{2}{*}{pain}                               & unknown or naturally occurring                         & face                                                     & +                                          & \checkmark               &                          \\
Pessanha et al.~\cite{pessanha2020towards}          &                                                       &                                                     & unknown or naturally occurring                         & face                                                     & +                                          & \checkmark               &                          \\ \midrule
Lencioni et al.~\cite{lencioni2021pain}             & \multirow{8}{*}{Horses}                               & \multirow{6}{*}{pain}                               & surgical castration                                    & face                                                     & +                                          & \checkmark               &                          \\
Hummel et al.~\cite{hummel2020automatic}            &                                                       &                                                     & unknown or induced pain                                & face                                                     & +                                          & \checkmark               &                          \\
Broom\'e et al.~\cite{broome2019dynamics}           &                                                       &                                                     & induced pain                                           & body and face                                            & +                                          &                          & \checkmark               \\
Broom\'e et al.~\cite{broome2022sharingpain}        &                                                       &                                                     & induced pain                                           & body and face                                            & +                                          &                          & \checkmark               \\
Rashid et al.~\cite{rashid2022equine}               &                                                       &                                                     & induced pain                                           & body                                                     & +                                          &                          & \checkmark               \\
Reulke et al.~\cite{reulke2018analysis}             &                                                       &                                                     & vet. procedure                                         & body                                                     & -                    &                          & \checkmark               \\
Corujo et al.~\cite{corujo2021emotion}              &                                                       & emotion                                             & unknown                                                & body and face                                            & +                                          & \checkmark               &                          \\
Li et al.~\cite{li2021automated}                    &                                                       & -                                                   & -                                                      & face                                                     & -                                          & -                        & -                        \\ \midrule
Feightelstein et al.~\cite{feighel}                 & Cats                                                  & pain                                                & vet. procedure                                         & face                                                     & +                                          &                          & \checkmark               \\ \midrule
Morozov et al.~\cite{morozov2021automatic}          & \multirow{2}{*}{Macaques}                             & \multirow{2}{*}{emotion}                           & induced behavior                                       & face                                                     & -                & \checkmark               &                          \\
Blumrosen et al.~\cite{blumrosen2017towards}        &                                                       &                                                     & induced behavior                                       & face                                                     & -                & \checkmark               &                          \\ \midrule
Zhu~\cite{zhu2022dogpain}                           & \multirow{4}{*}{Dogs}                                 & pain                                                & naturally occuring                                     & body                                                     & +                                          &                          & \checkmark               \\
Franzoni et al.~\cite{franzoni2019preliminary}      &                                                       & \multirow{3}{*}{emotion}                            & unknown                                                & face                                                     & +                                          & \checkmark               &                          \\
Boneh-Shitrit et al.~\cite{tali}                    &                                                       &                                                     & induced behavior                                       & face                                                     & +                                          & \checkmark               & \checkmark               \\
Ferres et al.~\cite{ferres2022predicting}           &                                                       &                                                     & unknown                                                & body                                                     & +                                          &                          & \checkmark               \\ \midrule
Statham et al.~\cite{statham2020quantifying}        & Pigs                                                  & emotion                                             & induced behavior                                       & body                                                     & -                                          &                          & \checkmark               \\ \bottomrule
\end{tabular}}
\end{sidewaystable}

To systematize the existing body of research, % on computer vision based approaches of recognizing animal pain and affective states
in this section, we review and analyze twenty state-of-the-art works addressing assessment, classification and analysis of animal emotions and pain.
The works are presented in Table \ref{tab:works}, and are classified according to the following characteristics: 

\begin{itemize}
 \item \textit{Species:} We restrict the scope of our review to mammals. The list of included species can be seen in Table \ref{tab:works}. \\   % larger than, and including, rodents. % Whenever we refer to mammals, throughout the article, we exclude humans. 

 \item \textit{State and state classifier:} Our main aim is to cover works focusing on recognition of internal states in animals, falling under the categorization of emotion and pain. Thus, the main type of works we are interested in are those providing a classification method (%pain/no pain
 {pain score}, or classifying specific emotional states). In the table's column `State classifier', this is signified by `+'. There are five works in the table marked with `-', which do not provide such a classifier, but nevertheless develop computer vision-based methods explicitly designed to investigate behavior patterns related to affective or pain states.\\
 %which are deemed as useful or relevant for such task.  
 
 %Although both emotions and pain can be thought of internal states, historically they have been investigated separately due to the complex differences between them; therefore, we differentiate between pain and emotion recognition. 

 \item \textit{Stimulus:} This category designates the type of stimulus that the animals have been subject to during data collection.\\

 \item \textit{Focus area:} We restrict our attention here to facial and bodily behavior indicators, or a combination of the two. \\

 \item \textit{State annotations:} This column is divided into two options: behavior-based or stimulus-based state annotations, respectively. The behavior-based annotations are purely based on the observed behaviors, without regard to when the stimulus (if there was one) occurred. For stimulus-based annotations, the ground-truth is based on whether the data was recorded during an ongoing stimulus or not.

\end{itemize}

In this section, the meta-analysis of the works in Table \ref{tab:works} is organized according to the different stages of a typical workflow in studies within this domain: data collection and annotation, followed by data analysis (typically, model training and inference) and last, performance evaluation. For each of these stages, we classify the methods and techniques applied in these works, highlight commonalities and discuss their characteristics, limitations and challenges.

\subsection{Data Collection and Annotation}\label{sec:data}

Hummel et al. \cite{hummel2020automatic} highlight some important challenges in addressing automated pain recognition in animals, most of which can also be generalized to recognition of affective states.
The first challenge is the lack of available datasets, compared to the vast amount of databases in the human domain \cite{Hassan2021AutomaticDO}. %(see, e.g. \cite{owusu2021facial} for a comprehensive list). 
This is due to the obvious difficulties of data collection outside of laboratory settings, especially for larger animals -- companion as well as farm animals.  
Secondly, particularly in the case of domesticated species selected for their aesthetic features, there may be much greater variation in facial texture and morphology than in %, compared to 
humans. This makes population-level assessments difficult, due to the potential for pain-like facial features to be more present/absent in certain breeds at baseline~\cite{finka2019geometric}. Finally, and perhaps most crucially: there is no verbal basis for establishing a ground truth, whereas in humans, self-reporting is commonly used. This complicates data collection protocols for non-human animals, sometimes requiring conditions where the induction of a particular affective state and its intensity must be closely controlled and regulated, and/or requires rating by human experts, potentially introducing biases. %and a circular logical flaw relating to human expectation for objective validity. 
Below we examine the data collection choices and data annotation methods in the works reviewed here.

\subsubsection{Data Collection}
%Mention how the different works collected their data
{\bf Recording equipment.}
Choosing the equipment with which to record visual data is the first step to acquire data. Since this survey concerns vision-based applications, this would either be an RGB, depth or infrared camera. As stated in Andersen et al. \cite{andersen2021towards}, the requirement on resolution for machine learning applications is often not a limiting factor. Some of the most frequently used deep neural network approaches work well with inputs of approximately 200x200 pixels. However, subtle cues, such as muscle contractions, can be difficult to detect reliably in low-resolution images. Infrared cameras are typically used to be able to monitor behavior during night, in order not to disturb the sleep cycle of the animal with artificial light. % \sbr{Maybe Recording equipment could be removed actually. }
While clinical annotations are typically done by veterinarians, animal images scraped from the Internet without any expert annotations can also be useful for training computer vision tools. Large object recognition datasets such as MS COCO include animal classes, although from a limited number of species \cite{Lin2014MicrosoftCC}. Therefore, models trained with those datasets can be helpful for detecting animals and for pose estimation. \\

\noindent {\bf Environment.}
%camera placing. laboratory box (mice) / farm/ stables/ vet clinic / ....number of angles. light control. 
Rodents are typically recorded in observation cages with clear walls to permit their recording~\cite{tuttle2018deep}. The camera is static and can cover the entire cage, but for facial analysis, only frames that have the rodent in frontal view are selected and used. Infrared cameras placed on top of the face are also used, but these observe movement patterns, and not facial expressions. Equines are recorded in a box \cite{rashid2022equine, Ask2020IdentificationOB} from multi-view surveillance cameras, or in open areas, but with static cameras placed at a distance to capture the animal from the side~\cite{gleerup2015equine,hummel2020automatic, broome2019dynamics, broome2022sharingpain}, or frontally, when the animal is next to a feeder~\cite{lencioni2021pain}. The side view makes observing the bodily behavior easier, but only one side of the face is visible. The presence of a neck collar or bridle is common for recordings, as the animals are often constrained. Cattle and sheep are recorded outdoors, in farms, with widely varying background and pose conditions~\cite{mahmoud2018estimation}. Recordings of animals from veterinary clinics, on the other hand, uses static cameras indoors, where the animal can move freely in a room~\cite{zhu2022dogpain}. This allows the expert to evaluate behavioral cues during movement. \ \\

\noindent{\bf Participants.} Controlling the data for specific characteristics of participants can lead to increased performance and its better understanding. On the other hand, generalizability of such models can be limited. Some studies reviewed here practiced control for color  
%\cite{tuttle2018deep} (white mice), \cite{andresen2020towards} (black mice), \cite{statham2020quantifying} (white pigs);  breed: \cite{feighel} (British Short Haired cats), \cite{tali} (Labrador Retriever dogs); and sex: \cite{feighel} (female cats), \cite{lencioni2021pain} (male horses). 
(e.g. white mice~\cite{tuttle2018deep}, black mice~\cite{andresen2020towards}, white pigs~\cite{statham2020quantifying}), breed (e.g. British Short Haired cats~\cite{feighel}, Labrador Retriever dogs~\cite{tali}), and sex (e.g. female cats~\cite{feighel}, male horses~\cite{lencioni2021pain}).  

% Equipment
% Multi-view

% ethics, characteristics, limitations
% intra-annotator agreement, potential issues and biases, setup angle, modality, how many cameras
% we'll see if we dissect the papers different datasets here or not

\subsubsection{Data Annotation}
\label{sec:annotation}

In the human domain, self-reporting is considered one of the most unobtrusive and non-invasive methods for establishing ground truth in pain \cite{labus2003self} and emotion research \cite{barrett2004feelings}. Furthermore, in emotion research the use of actors portraying emotions is a common method for data collection and annotation \cite{seuss2019emotion}. For obvious reasons, these methods are not usable for animals, making the establishment of ground truth with respect to their internal states {highly} challenging, and adding further complications to the data annotation stages. 

One possible strategy for establishing the ground truth can be based on designing or timing the experimental setup to induce the affective state or pain. In the case of pain, designing can refer to experimental induction of clinical short term reversible moderate pain using models known from human volunteers. In \cite{broome2019dynamics}, e.g., two ethically regulated methods for experimental pain induction were used: a blood pressure cuff placed around one of the forelimbs of horses, or the application of capsaicin (chili extract) on the skin of the horse.  Another possibility is to time data collection after %, e.g., in the context of 
a clinical procedure. This is the case in \cite{feighel}, where female cats undergoing ovariohysterectomy were recorded at different time points pre- and post-surgery.
    
In the case of affective states, state induction can be performed, e.g.,  using triggering stimuli and training to induce emotional responses of different valence. For instance, in \cite{tali} the data was recorded using a protocol provided in \cite{bremhorst2019differences}, using a high-value food reward used as the triggering stimulus in two conditions – a positive condition predicted to induce positive anticipation, and a negative condition predicted to induce frustration in dogs. In Lundblad et al. \cite{lundblad2021effect}, stress was induced by letting out one out of two horses that were normally let out together (herd mates). After between 15 and 30 minutes alone, the horse that was not let out showed a marked stress response. 
The presence of people (such as the owner of pets) can influence the behavior of the animal, and should be taken into account in analyses. In cases when no control over the state of the animal is exercised, the animal can be recorded in a naturalistic setting, such as farms \cite{mahmoud2018estimation,li2021automated} or stables, or even laboratory cages. Data collected from veterinary clinics with ``naturally occurring" pain denotes animals brought to the clinic under pain, as opposed to ``induced" pain, which is a more controlled setting.

   Cases when data is scraped from the Internet \cite{franzoni2019preliminary,ferres2022predicting} should be treated with caution in this context (and are accordingly labeled as `unknown' in Table \ref{tab:works}), as the degree of control cannot be asserted. When the state is not controlled, the only available option for establishing ground truth is by human annotators. This may introduce bias and error, depending on the annotators' expertise (veterinarians, behavior specialists, laymen), the number of annotators and agreement between them, and also on whether specific measurement are instruments used (e.g., validated grimace scales \cite{dalla2014development}).

Table \ref{tab:works} includes a classification of the works reviewed here according the data annotation strategies discussed above. %We only include here works that present classifiers for pain/emotion (thus, excluding here \cite{reulke2018analysis,morozov2021automatic,blumrosen2017towards,statham2020quantifying}). 
In the studies of \cite{tuttle2018deep,andresen2020towards,feighel}, the animal participants underwent a surgical procedure. In \cite{tuttle2018deep,andresen2020towards}, the obtained images were then rated by human experts based on the mouse grimace scale. In \cite{feighel}, on the other hand, the images were taken at a point in time where the presence of pain was reasonable to assume (between 30-60 min. after the end of surgery, and prior to administration of additional analgesics). %Since the facial landmarks in this data were annotated by a human expert, this is signified by `-(+exp)' in the table. 
In \cite{broome2019dynamics,broome2022sharingpain,rashid2022equine} experimental pain is induced using controlled procedures for moderate and reversible pain. The dataset used in \cite{hummel2020automatic} is composed from several sources: a clinical study, where pain was experimentally induced, images taken at a home housing older horses, and images provided by horse owners. In \cite{corujo2021emotion}, the data is collected from different private sources where the horse and context of the photo was familiar, guiding annotation. However, the state annotation was performed by laymen. Since the states or contexts are not described for this dataset, we have marked this as `unknown' in Table \ref{tab:works}. %This is signified by `mixed' in the table.
\cite{franzoni2019preliminary,ferres2022predicting} use images scraped from the web, thus the state control is stated `unknown'. \cite{tali} uses images of dogs taken in an experiment where emotional states are induced by food rewards, with no human involved in the annotation loop.

Induced approaches, if performed properly and in a controlled and reproducible manner, have the potential to reduce human bias and error, while the use of data from unknown sources can be problematic in terms of bias, error and noise in ground truth annotation \cite{waran2010recognition,price2002pilot}.

\subsection{Data Analysis}
\label{sec:analysis}

The stage of data analysis typically involves
developing a data processing pipeline, the input of which is  images or videos, and the output of which is a classification of an affective state (usually class of emotion), or pain classification (binary yes/no) or degree assessment (more than two classes). The pipeline may involve one or more steps, and address body/face as a whole, or process first their specific parts.

\subsubsection{Input: frames vs. sequences of frames}

Computer vision-based methods operate on data in the form of images or image sequences (videos). 
This implies the following three main modes of operation with respect to temporality:

\begin{itemize}
    \item \textit{Single frame basis.} This %simplest 
    route is taken by the majority of works reviewed {in the survey}. %here
    This {is the simplest and least expensive} option 
    in terms of computational resources.  
    
    \item \textit{Frame aggregation.} Using frame-wise features, some works address classification of videos by aggregating the results of classifiers working with single frames, thus at least partially incorporating information contained in sequences of frames. This is the route taken in \cite{tuttle2018deep} and  \cite{pessanha2020towards}. 
    
    \item \textit{Using spatio-temporal representations.} A third route is to learn spatiotemporal features from video given as sequential input to a deep network. This is done in \cite{broome2019dynamics, broome2022sharingpain,zhu2022dogpain}, and enables the detection of behavioral patterns that extend over time. Apart from presenting computationally heavy training, this method requires more data than frame-wise approaches. On the other hand, having access to video recordings often is synonymous to having access to a lot of data. However, this is relative, and the horse video datasets used in \cite{broome2019dynamics, broome2022sharingpain}, which have a duration of around ten hours each, are comparable in scale to older well-known video datasets such as UCF-101 \cite{ucf101} (30h), but not to newer ones, such as Kinetics \cite{KineticsDataset} (400h). % \az{Sofia - can you please help discuss your dynamics work}

\end{itemize}

As was found in \cite{broome2019dynamics}, for the case of horse pain detection, temporal information is crucial for discriminating pain. Temporal information has also previously been found to improve recognition of human pain \cite{Bartlett2014AutomaticDO}.  In \cite{rashid2022equine}, the use of single-frame and sequential inputs for pain classification are also compared, in a multiple-instance learning (MIL) setting. MIL can be seen as lying somewhere in between temporal aggregation and spatiotemporal features, in being a more advanced form of temporal aggregation, within a learning framework. Using single frames gives more control and promotes explainability, but leads to information loss. Working with video input, on the other hand, rather than single-frame input, is costly.

Thus, choosing the mode of operation is ultimately goal dependent. If the goal is to count in how many frames in a certain video segment that a horse has its ears forward (an estimate of the fraction of time with ears kept forward), it suffices to detect forward ears separately for each frame, and subsequently aggregate the detections across the time span of interest. If, on the other hand, the goal is to study motion patterns of the horse, or distinguish between blinks and half-blinks for an animal, it is crucial to model the video segment spatiotemporally. An explorative search for behaviors which potentially extends over time might also be desirable, and the degrees of freedom offered by spatiotemporal feature learning approaches is useful for such a task.

\begin{table}[tbh]
\caption{An overview of the approaches taken at the analysis stage, categorized according to whether the methods are parts-based or holistic, based on frame-wise or video information, and whether the features are learned or hand-crafted.}
\adjustbox{max width=\textwidth}{ \begin{tabular}{llllll}
Study & Species & Part/Holistic & Input & Features \\ \hline
%Sotocinal et al.~\cite{sotocina2011rat}    &  mice       &   pain  &          &  \\
Tuttle et al.~\cite{tuttle2018deep}    &   mice     &  holistic    &          frame & learned   \\
Andresen et al.~\cite{andresen2020towards}    &  mice       &  holistic    &          frame &  learned\\
Mahmoud et al.~\cite{mahmoud2018estimation}    &     sheep    &    parts-based   &          frame &  hand-crafted (low-level)\\

Pessanha et al.~\cite{pessanha2020towards}    &     sheep    &    parts-based   &          frame(ag) & hand-crafted (low \& high-level)\\

 Lencioni et al.~\cite{lencioni2021pain}   &  horses       &  parts-based     &          frame & learned \\
Hummel et al.~\cite{hummel2020automatic}    &  horses       &  parts-based    &          frame & hand-crafted (low-level)\\

Broom\'e et al.~\cite{broome2019dynamics}    &  horses       &   holistic    &          video & learned\\
Rashid et al.~\cite{rashid2022equine}    &  horses       &   holistic    &         video & learned\\

Reulke et al.~\cite{reulke2018analysis}    &  horses       &   holistic    &          video& -\\

Corujo et al.~\cite{corujo2021emotion}    &  horses       &   holistic    &           frame & learned\\
Li et al.~\cite{li2021automated}    &   horses       &  parts-based                & frame & learned\\

 Feightelstein et al.~\cite{feighel} 1   &  cats        &   holistic    &          frame & learned\\
  Feightelstein et al.~\cite{feighel} 2   &  cats        &   holistic    &          frame & hand-crafted (high-level)\\
Morozov et al.~\cite{morozov2021automatic}    &    macaques     &   holistic             &  frame & hand-crafted (high-level)\\
Blumrosen et al.~\cite{blumrosen2017towards}    &   macaques       &  holistic                & frame& hand-crafted (high-level)\\
Zhu~\cite{zhu2022dogpain}    &  dogs           & holistic & frame & mixed\\
Franzoni et al.~\cite{franzoni2019preliminary}    &   dogs      &    holistic   &          frame &learned   \\
Boneh-Shitrit et al.~\cite{tali}    &  dogs       &  holistic     &  frame & learned       \\
Ferres et al.~\cite{ferres2022predicting}    &  dogs       &  emotion     & frame & hand-crafted (high-level)\\

Statham et al.~\cite{statham2020quantifying}    &  pigs       &  emotion     & frame & -\\
   &        &  
   & 
\end{tabular}
} % adjustbox closing bracket
\label{table:bigtable}
\end{table}

\subsection{Parts-based vs. Holistic Methods}
Methods for computer vision-based human facial analysis are commonly divided into local parts-based and holistic methods, differing in the way facial information is processed \cite{wang2018facial,wu2019facial}. Parts-based methods divide the input data
%face
into different areas, e.g., considering different facial features separately, while holistic methods process the information of the input data as a whole, be it at the body or face level.

The idea of dividing the face into regions, or parts{,} is especially relevant for works on pain assessment that are based on species-specific grimace scales. Such scales typically divide the animal face into at least three parts, including ears, eyes and nose/mouth/nostrils. 
One example is the work of Lu et al. \cite{lu2017estimating}, providing a multi-level pipeline for assessment of pain level in sheep, based on the sheep facial expression pain scale (SPFES \cite{mclennan2016development}), according to which the sheep face is divided into regions of eyes, ears and nose. Although the cheek and lip profile are also
discussed in the SPFES, they are omitted in \cite{lu2017estimating}, because the sheep dataset in question only contains frontal faces, and these features
can hardly be seen on a frontal face. The eyes and ears are further split into right and left regions each. Each of these regions correspond to one out of three action units (AUs) defined based on the SPFES taxonomy {(pain not present (0), pain moderately present (1), or pain present (2))}. For instance, the ear region can correspond to one of the following AUs: ear flat (pain level=0), ear rotated (pain level=1) and ear flipped (pain level=2). SVM classifiers predicting the pain level for each of the five regions were then trained separately on each facial feature, using Histogram of Oriented Gradients (HOG), to depict the shape and texture of each feature. To aggregate these results,
the scores for symmetric features (eyes, ears) were averaged, and 
all three feature-wise scores (ear, eye, nose) were averaged again to obtain the overall
pain score. 

Another example of {a} parts-based approach is provided in \cite{lencioni2021pain} in the context of horse pain. Based on the horse grimace scale \cite{dalla2014development}, this work also focuses on three regions of the horse face: ears, eyes, and mouth and nostrils, training 
three separate pain classifier models based on convolutional neural networks (CNNs) for each of the regions. The outputs of these models are then fused using a fully connected network for an overall pain classification.  A parts-based approach to AU recognition is presented in Li et al. \cite{li2021automated}, where each AU is recognized on cropped image regions specific to the AU in question. Their results show that such close-up crops of the eye-region or lower-face region are necessary for the performance of the classification in their framework. 

In general, as the field of pain and affective state recognition is only beginning to emerge, using a parts-based approach can provide important insights on the role of each of the facial regions in pain expression. Interestingly, the results of \cite{lencioni2021pain} %and \cite{li2021automated}
indicate that ears provide better indication for pain level in sheep and horses than the other regions (although this should be considered with caution due to the imbalance of the dataset in terms of different parts, see also discussion in Section \ref{sec:best}). %Ears are also mentioned in \cite{li2021automated} as an important characteristic for pain behavior, but these action descriptors are not included in the article's results since they required temporal modeling and this was out of the scope of the work. 
Further exploration of parts-based approaches in additional species can provide insights into the importance of the regions, and thus allow methods to fine-tune the aggregation of a general pain score in future studies. The column `Part/Holistic' in Table~\ref{table:bigtable} classifies the works across the dimension of holistic vs. parts-based approaches.

\subsubsection{Hand-crafted vs. learned features}

A major focus in computer vision is to discover, understand, characterize, and improve the features that can be extracted from images. Traditional features used in the literature have been manually designed, or `hand-crafted'{,} overcoming specific issues like occlusions
and variations in scale and illumination, such as histograms of oriented gradients~\cite{nanni2017handcrafted}.  Traditional {computer vision} methods, prior to the deep learning era, have typically been based on hand-crafted features. The shift toward automatically learning the feature representations from the data occurred progressively during the 2010s as larger datasets were made public, GPU-computing became more accessible and neural network architectures were popularized in both the machine learning literature and in open-source Python frameworks, such as Tensorflow~\cite{tensorflow2015-whitepaper}. This new computing paradigm is commonly known as deep learning, where the word deep refers to the hierarchy of abstractions that are learned from data, and stored in the successive layer parameters~\cite{lecun2015deeplearning}.

The above context has important implications in the context of our domain. The first implication is related to dataset size: methods using hand-crafted features can be applied to small datasets, whereas deep learning methods require larger amounts of data. The second implication is the explainability of the approaches: hand-crafted features allows for a clearer understanding of the inner workings of the method, while learned features lead to `black-box' reasoning, which may be less appropriate for clinical and welfare applications, such as pain assessment. % in which{, for example,} automated recognition of pain is envisioned. 

Hand-crafted features can exist on multiple levels, which we roughly divide into two: 
{\em low-level} features are technical and may consist of pre-defined notions of image statistics (such as histograms of oriented gradients, or pixel intensity in different patches of the image). {\em High-level} features, in our context, are semantically grounded, typically based on species-specific anatomical facial and/or body structure, grimace scales or action units. As these features promote explainability, we refer to them as intermediate representations; these will be discussed in more detail further down.

The column `Features' in Table \ref{table:bigtable} classifies the works across the dimension of learned vs. hand-crafted features. The types of high-level features used in \cite{pessanha2020towards,feighel,ferres2022predicting,morozov2021automatic,blumrosen2017towards}
are further discussed in Section \ref{sec:reps}.

\subsection{Increasing Explainability: Intermediate Representations}
\label{sec:reps}

Higher-level features are features that have semantic relations to the domain of affective states and pain, e.g., through facial or bodily landmarks, grimace scale elements, action units, or pose representations. As such, these are highly valuable for the explainability of the different classification methods. These features are usually used in computational pipelines involving a number of pre-processing steps. They can be built either manually, or using classifiers based either on lower-level
hand-crafted or learned features. Below we discuss some important types of intermediate representations used in the works surveyed here, and how they are computed and used: 

\begin{itemize}
    \item {\em Facial Action {Units}}. Morozov et al. \cite{morozov2021automatic} and Blumrosen et al. \cite{blumrosen2017towards} apply two different approaches to address the recognition of facial actions in macaques as an intermediate step towards automated analysis of affective states. \cite{morozov2021automatic} addresses six dominant action units from macaque FACS {(MaqFACS \cite{parr2010brief})}, selected based on their frequency and importance for affective communication, training a classifier on data annotated by human experts. \cite{blumrosen2017towards} addresses four basic facial actions: neutral, lip smacking, chewing and random mouth opening, using an unsupervised learning approach without the need for annotation of data. Both works use eigenfaces \cite{donato1999classifying} as hand-crafted lower level features, an approach which uses PCA analysis to represent the statistical features of facial images. Lu et al. \cite{lu2017estimating} provide a pipeline for pain level estimation in sheep, in which automated recognition of nine sheep facial action units is performed using classifiers based on histograms of gradients as lower level hand-crafted features. 
    The AUs are related to SPFES, a standardised sheep facial expression pain scale  \cite{mclennan2016development}.

     \item {\em Landmarks/Keypoints}. 
     One of the approaches investigated in~\cite{feighel} in the context of cat pain is based on facial landmarks{, } specifically chosen for their relationship with underlying musculature, and relevance to cat-specific facial action units (CatFACS). The annotation of the 48 landmarks was done manually. {In the pain recognition pipeline, these landmarks are } %, and they were then 
     transformed into multi-region vectors~\cite{8997580}, which are then fed to a multi-layer perceptron neural network (MLP). 
     
     \quad The approach of~\cite{ferres2022predicting} for dog emotion recognition from body posture  uses 23 landmarks on both body and face. 
     The landmarks are automatically detected by a model based on the 
     DeepLabCut framework \cite{mathis2018deeplabcut}, and trained on existing datasets of landmarks \cite{cao2019cross,biggs2020left} containing subsets of the 23 landmarks. Two approaches are then examined for emotion classification: 1) feeding the raw landmarks to a neural network, and 
     2) computing body metrics introduced by the authors and feeding it to simpler decision tree classifiers to promote explainability. For the decision tree approach, the authors use a variety of body metrics, such as 
     body weight distribution, and tail angle. The former is calculated using the slope of the dorsal line, which is a hypothetical line between the withers keypoint and the base of the tail keypoint, and the latter by the angle between the dorsal line, and the hypothetical line between the base of the tail and the tip of the tail. 
     
     \item  {\em Pose representations.} In  \cite{rashid2022equine}, multi-view surveillance video footage is used for extracting a disentangled horse pose latent representation. This is achieved through novel-view synthesis, i.e., the task of generating a frame from viewpoint $j$, given a frame from viewpoint $i$. The latent pose arises from a bottleneck in an encoder-decoder architecture, which is geometrically constrained to comply with the different rotation matrices between different viewpoints. The representation is useful in that it separates the horse from its appearance and background, to remove any extraneous cues for the task which may lead to overfitting. The representation is subsequently fed to a horse pain classifier. The pain classification is cast as a multiple instance learning problem, on the level of videos. In \cite{zhu2022dogpain}, a pose stream is combined with a raw RGB stream in a recurrent two-stream model to recognize dog pain, building on the architectures used in \cite{broome2019dynamics, broome2022sharingpain}. This constitutes an interesting example of mixing intermediate with fully deep representations. % (as opposed to the more standard approach of single frames). 

\end{itemize}

\subsection{Going Deep: Black-Box Approaches}

{As noted above,} deep learning approaches are becoming increasingly popular in the domain of human affective computing as they require less annotation efforts if transfer learning is leveraged, and no efforts for hand-crafting features. Yet, the resulting models provide what is called `black-box' reasoning, which does not lend itself easily for explaining the classification decisions in human-understandable terms (see, e.g., \cite{london2019artificial}). This is a crucial aspect, especially in the context of clinical applications and animal welfare. 

The convolutional neural network (CNN) is the most popular type of deep model used in the works surveyed in this article. Examples of used CNN architectures include ResNet50 \cite{corujo2021emotion,feighel,tali,andresen2020towards}, InceptionV3 \cite{tuttle2018deep,andresen2020towards} and AlexNet \cite{franzoni2019preliminary}.  One work addressing dog emotion \cite{tali} compared a CNN (ResNet50) to a Vision Transformer, ViT \cite{dosovitskiy20vit}, a model fully based on attention mechanisms instead of convolutions, finding the latter to perform better. The authors hypothesize that this is due to
the sensitivity of such models to object parts \cite{amir2021deep}, and suggest that automated emotion
classification requires understanding at the object-part level.

Another type of neural network is the deep recurrent video model used in \cite{broome2019dynamics, broome2022sharingpain, zhu2022dogpain}, based on the ConvLSTM \cite{Shi2015ConvolutionalLN} layer. A ConvLSTM unit replaces matrix multiplication by convolution in the LSTM equations, thus allowing for spatial input rather than 1D vectors in a recurrent setting. In this way, spatial and temporal features can be learned simultaneously, instead down-sampling the spatial features prior to temporal modeling. The best performing version of the model in \cite{broome2019dynamics} takes both RGB and optical flow input in two separate streams with late fusion. In \cite{broome2019dynamics}, this model is compared to a frame-wise InceptionV3 and to a VGG \cite{Simonyan2015VeryDC} network with a standard LSTM layer on top, thus taking sequential input. Even if the VGG+LSTM obtains numerical results not far from the two-stream ConvLSTM, qualitative examples using Grad-CAM \cite{GradcamSelvarajuCDVPB17} indicate that the ConvLSTM learns more relevant features. In \cite{broome2022sharingpain}, it is also found that an I3D model \cite{Carreira2017QuoVA} (a deep 3D convolutional neural network) can learn spatiotemporal features for pain recognition, but that it performs weaker in terms of generalization to a new pain type compared to the ConvLSTM model. It is hypothesized that this overfitting behavior of the I3D is due to its large parameter count (around 23M) relative to the ConvLSTM (around 1M), and that smaller video models may be advantageous for this type of fine-grained classification task, where motion cues should matter more than appearance and background of the videos. In~\cite{zhu2022dogpain}, LSTM and ConvLSTM layers are used in a dual-branch architecture, where one branch processes keypoint-based representations, and the other RGB-based representations.
%...\az{Sofia - please help here with LSTM models}

\subsection{Performance Evaluation}
\label{sec:evaluation}

Understanding and scrutinizing the methods for measuring performance are key when comparing approaches in recognition of affective states and pain. 
In this section, we give an overview of the evaluation protocols as well as classification performances of the different approaches listed in Table \ref{tab:works}. 
We emphasize that comparing the performance of classifiers of affective states and pain in animals presents great challenges, and cannot be done solely on the basis of the numbers as measured by performance metrics. This is due to the significant differences in 
data
acquisition (different lighting conditions, camera equipment, recording angles), as well as in ground truth annotation (naturalistic vs. controlled setting, induced vs. natural emotional state/pain, degree of agreement between annotators and their expertise). Even when all of these factors are comparable, technical choices such as differences in data balance or validation method greatly affect performance metrics (we discuss these aspects in the next section and provide some best practice recommendations on the basis of the analyzed works). 

Table \ref{table:eval} dissects the results and evaluation protocols of the works surveyed here. However, we have excluded works which do not involve a down-stream classification task, but rather describe pain behavior using computer vision (e.g., Reuss et al. \cite{ruess2019equine} and Statham et al. \cite{statham2020quantifying}), as well as pre-prints, and the three works which address only AU classification. The categories included in Table \ref{table:eval} are explained as follows.
%The AU classification articles are difficult to compare in a table, but will still be discussed further down in this section.

%(with the exception of works which do not present an end-to-end classifier, or are yet unpublished), split into pain and emotions using the following dimensions:
\begin{itemize}
    \item \textit{Species:} As discussed in Section \ref{sec:data}, it is important to consider the data collection protocols, for variations in  conditions of lightning, angle of recording, etc. (e.g., for small laboratory animals such as mice, compared to larger animals such as sheep and horses). Also, differences across breeds, age, color and sex may be an important factor.
    
    \item \textit{CrossVal:} the method used for cross-validation. We use the following abbreviations: single train-test-validation split (STTVS) (as opposed to k-fold cross-validation) and leave-one-animal-out (LOAO). 
    
    \item \textit{SubSep:} whether subject separation (subject exclusivity) was enforced in the splitting between train, test and validation sets. 
    
    \item \textit{SepVal:} whether the validation set was different than the test set. 
    In general, the test set should be fully held-out, ideally until all the experiments are finished. Since this often is difficult to achieve because of data scarcity, it is good practice to base model selection on a validation set, to then evaluate the trained model on the held-out test set. 
    %Four out of thirteen works do not use separate validation and test sets, roughly a third from both the pain and emotion categories. 
    
    \item \textit{\#cl:} number of classes used for classification. In emotion recognition, the classes correspond to the emotions studied (e.g., relaxed/curious in \cite{corujo2021emotion}, or happy in \cite{ferres2022predicting,franzoni2019preliminary} ). In pain recognition, there is binary (pain/no pain) or three-level classification. The number of classes is important, as methods using different number and types of classes are often incomparable in terms of performance. In such cases, e.g., multi-class classifications (such as degree-based classification of pain, e.g., the ternary classification in \cite{lencioni2021pain}), can be collapsed to binary classification (pain/no pain) to allow for a comparison.   
    
    \item \textit{BM:} data balancing method used. Data imbalance significantly affects performance metrics, thus using balancing methods in cases of greatly imbalanced datasets is important. We further elaborate on this point in the next section. 
    
    \item \textit{Metrics of accuracy, precision, recall and F$_1$.} Whenever confusion matrices are provided in the articles, we completed the computations for metrics for precision and recall, if not already given. Without published confusion matrices, we can only rely on the numbers in the articles. Hence, the measures that we could not obtain were simply left as blank (-) in the table.

\end{itemize}

It should be noted that the practice of  subject-exclusive evaluation is important. By separating the subjects used for training, validation and testing respectively, generalization to unseen subjects is enforced, making sure that no specific features of an individual are used for classification. Subject-exclusive evaluation is trivially guaranteed in cases when there is one sample per subject, as is often the case with data scraped from the web, e.g., \cite{franzoni2019preliminary,hummel2020automatic}. In datasets of this type, however, there is greater risk for bias and noise, introduced by data collected under unknown conditions. Datasets from private sources \cite{hummel2020automatic, corujo2021emotion} or from the authors own clinical trials \cite{tuttle2018deep, lencioni2021pain}, on the other hand, typically involve a smaller number of individual animals, meaning that the risk is higher for the same animal with a similar expression to be present both in the training and testing set. In such cases the cross-validation method leave-one-animal-out is highly recommended, which also naturally enforces subject-exclusivity. The latter can also be exercised with other cross-validation methods, such as STTVS. We further elaborate on best practices in the context of cross-validation in the next section.

%Our motivation here is to shed light on how much variation there can be between different evaluation approaches, and to recommend best practices for reliable evaluation that avoids the common pitfall of overfitting. % \az{perhaps a ref?}.
%We compare the protocols using the dimensions described below.

%The most commonly used metrics in classification problems are accuracy, precision, recall and $F_1$-score. %\az{PERHAPS DEFINE THEM? or at least give refs} \sbr{I think I feel that we don't need to define them for a CV-audience (which is IJCV).}
%When we present precision and recall scores in Table \ref{table:eval}, these are the average for the precision and recall for each of the separate classes (including when there are only two-classes).  %Accuracy is the total number of correctly classified samples out of the total number of classifications made. %Precision and recall are typically defined for a binary scenario, of positive and negative samples. In that case, one can define true and false negatives, as well as true and fal

%Another important dimension in evaluation protocols is validation. 
%Perhaps add more details here on what validation is, what cross-validation is, and then what is subject-exclusiveness and separate validation set.}

% \usepackage{booktabs}
% \usepackage{multirow}
\begin{sidewaystable*}[]
\small
\caption{Overview of the performance of the published works that do down-stream classification in either pain or affective states.}
\adjustbox{max totalheight=\textheight}{
\begin{tabularx}{\textwidth}{@{}lllccrlllll@{}}
                            & \textbf{Species} & \textbf{CrossVal}      & \textbf{SubSep} & \textbf{SepVal}   & \multicolumn{1}{l}{\textbf{\# cl}} & \textbf{Balancing} & \textbf{Acc.} & \textbf{P} & \textbf{R} & \textbf{F1} \\
                            \toprule
\textbf{Pain}               &                        &                            &                      & \multicolumn{1}{l}{}                 &                                       &               &                         &                      &             \\ \midrule \\
Tuttle \cite{tuttle2018deep}                     & Mice &     STTVS            & No                         & Yes                  & 2                                    &  oversampling                                    & 93.2          & 93.7                    & 93.3                 & 93.5        \\
Andresen \cite{andresen2020towards}                   & Mice &   10-fold               & Yes                        & No                   & 2                                    &  -                                     & 89.8          & -                       & -                    & -           \\
Pessanha \cite{pessanha2020towards}                   & Sheep &  5-fold             & No                         & No                   & 2                                    &  -                                     & 78.0          & 83.0                    & 68.0                 & 73.0        \\
Lu  \cite{lu2017estimating}                        & Sheep &       10-fold          & No                         & No                   & 3                                    &     random undersampling                                  & 64.0         & 63.9                    & 59.8                 & 61.8        \\

Lencioni \cite{lencioni2021pain}                   & Horses & 10-fold                & No                         & Yes                   & 3                                    &   -                                   & 75.8         & 76.2                    & 75.8                 & 76.0       \\

%\multirow{2}{*}{Lencioni \cite{lencioni2021pain}} &  \multirow{2}{*}{Horse}  & \multirow{2}{*}{No}        & \multirow{2}{*}{Yes}  & 3                                    & \multirow{2}{*}{}                     & 75.8          & 76.2                    & 75.8                 & 76.0        \\
 %                           &                        &                            &                      & 2                                    &   approx. balanced                                    & 88.3          & 87.2                    & 86.3                 & 86.7        \\
%Hummel \cite{hummel2020automatic}                     & Horses & STTVS   & No                         & Yes                  & 3                                    &   -                                    & -             & -                       & -                    & 87.0        \\
Broomé \cite{broome2019dynamics}                & Horses  & LOAO               & Yes                        & Yes                  & 2                                    &   -                                    & 75.4          & -                       & -                    & 73.5        \\
Broomé \cite{broome2022sharingpain}                 & Horses  &    LOAO            & Yes                        & Yes                  & 2                                    &  oversampling clips w/ half stride                                     & -             & -                       & -                    & 58.2        \\
Rashid   \cite{rashid2022equine}                   & Horses &  LOAO               & Yes                        & Yes                  & 2                                    &   -                                    & 60.9          & -                       & -                    & 58.5        \\
Feighelstein \cite{feighel}               & Cats       &     LOAO        & Yes                        & Yes                  & 2                                    &  random undersampling                                     & 73.6          & 81.9                    & 70.1                 & 75.5        \\
                            &                        &                            &                      & \multicolumn{1}{l}{}                 &                                       &               &                         &                      &             \\
\textbf{Emotion}            &                        &                            &                      & \multicolumn{1}{l}{}                 &                                       &               &                         &                      &             \\ \midrule 
Corujo \cite{corujo2021emotion}                     & Horses &   5-fold              & No                         & Yes                  & 4                                    & balanced                                      & 65.0          & 60.0                    & 65.6                 & 62.7        \\
Franzoni \cite{franzoni2019preliminary}                   & Dogs &     5-fold               & No                         & No                   & 3                                    & -                                       & 95.3          & 93.3                    & 93.1                 & 93.1        \\
Ferres \cite{ferres2022predicting}                     & Dogs &      10-fold              & No                         & Yes                  & 4                                    & selected undersampling                                      & 67.5          & 68.4                    & 67.5                 & 67.9     \\ \bottomrule  
\end{tabularx}
} %adjustbox closing bracket
\label{table:eval}
\end{sidewaystable*}

\section{Best Practice Recommendations}\label{sec:best}
Based on the landscape of current state-of-the-art works reviewed in this survey, and learning from best practice recommendations from other scientific communities, such as human affective computing,  
we provide below some technical recommendations for best practices in future research on automated recognition of animal affective state and pain.

\subsection{Data Imbalance}

%The problem of data imbalance is commonly occurring in classification tasks within animal internal state recognition since the data where a particular state is controlled for is expensive to obtain and to annotate.
%frequently occurs in the context of classification tasks. %when the number of samples in one category (class) exceeds the amount in others.
In the field of affective computing for animals, and specifically in animal pain recognition, data imbalance problems are particularly acute, due to the difficulty to obtain samples of, e.g., the `pain' class, as opposed to the more available samples of the baseline.  Moreover, pain behavior is a complex concept to learn, for humans and non-humans. This poses a difficulty for learning algorithms, which may collapse and only predict the majority class, %due to created bias towards the majority group,
when in fact the minority class may carry important and useful knowledge. Both for classic machine learning methods and deep learning methods, the most common remedy is data-driven, where the relevant classes typically are over- or under-sampled \cite{kulkarni2020foundations, Buda2018ASS}.
%Therefore, when facing such disproportions, one must design an intelligent system that is able to overcome such a bias. 

The problem of learning from imbalanced data is well-studied in classical machine learning \cite{10.5555/1293951.1293954,thabtah2020data}.  A variety of methods to deal with data imbalance in this context have been proposed, see, e.g., \cite{kulkarni2020foundations} for a comprehensive list. One of the conclusions reached in \cite{10.5555/1293951.1293954} is that two essential factors which significantly impact performance are the degree of class imbalance and complexity of the concept to be learned.

In a deep learning context, data imbalance is often studied for datasets with a large number of classes and so called long-tail distributions of the minor classes \cite{Huang2016LearningDR, Li2020OvercomingCI, Cui2019ClassBalancedLB}, which is typically less relevant for our setting. However, \cite{Buda2018ASS} studies class imbalance in the context of CNNs on datasets with fewer classes, finding that oversampling does not necessarily contribute to overfitting.   

In \cite{Buda2018ASS}, it is stated that the most common approach for deep methods is oversampling. Modifying the loss function is another option in a deep learning setting \cite{Buda2018ASS}; this is commonly done for the above mentioned long-tail distribution scenarios. In random undersampling, instances from the negative class or majority class are selected at random, and removed until it matches the count of positive class or minority class, resulting in a balanced data set consisting of an equal number of positive and negative class examples. This method was used, e.g., in \cite{feighel}, addressing cat pain.

In the horse pain video dataset used in \cite{broome2019dynamics}, there is slight class imbalance (pain is the minority class, by around 40\%), when the sequences are extracted as back-to-back windows from the videos. No re-sampling is done in \cite{broome2019dynamics}, but an unweighted F1 average across the two classes is used to present more fair results than accuracy (since this metric requires performance on both the positive and negative class). The same class imbalance is addressed in a follow-up work \cite{broome2022sharingpain}, where video clips are over-sampled for the minority class. This is possible for video sequences, since one can easily adjust the stride of the extracted windows to obtain a larger number of sequences from the same video.

Another possibility is to use data augmentation, for instance by horizontally flipping images, adding noise to images, or randomly cropping images. However, this may change the distribution of the data, and should therefore be used with caution when only applied to one class, to avoid overfitting to an artificially augmented distribution. 
% if the paper is accepted, we can add Francisca's 3D-model based synthesis for data augmentation here. 

\begin{description} \item [Recommendation 1:] Data imbalance should be minimized using relevant data balancing techniques, such as oversampling, undersampling or loss modifications.
 \end{description}

 \subsection{Cross-Validation}   
 
 One crucial observation arising from our survey is that works in this domain typically
 use highly dimensional datasets (being computer vision-based), which commonly have a small number of samples because of the intrinsic difficulties with data collection involving animal participants. The combination of high dimensionality with a small number of participants (possibly with few repeated samples per participant) has a higher potential of leading to bias in performance estimation.

 As shown in Table \ref{table:eval}, the cross-validation techniques used in the surveyed papers include single train and test split, k-fold cross-validation (with k=5 or 10) and leave-one-animal-out methods. As previously mentioned, the latter means that the separation to training, validation and test sets is done on the basis of animal individuals, rather than on the basis of images or videos. 
 
 For deep methods, neural networks are typically trained throughout a number of epochs on a dataset, during which one can monitor the performance on a validation set after each epoch. This process allows you to choose the epoch where the model performs most optimally on the validation set. On the other hand, if the validation set is your final evaluation set, this amounts to adapting your model to your test set. Therefore, it is important to have a third split of the data -- the test set, on which you can evaluate your model, which has not been part of the model selection process.
 
 The training, validation and test splits can be constructed either randomly, or, ideally, in a subject-exclusive manner. In \cite{broome2019dynamics, broome2022sharingpain, rashid2022equine, feighel}, the presented results are averages of the test set results across a full test-subject rotation (each subject is used as test set once). Last, it can be mentioned that for deep learning methods, the random seed affects the initialization of the networks, when trained from scratch. If training has been carried out with different random seeds, it is important to present results that are averages of repeated such runs, to avoid cherry picking a particularly opportune training instance. Therefore, \cite{broome2019dynamics} additionally repeats each split five times, and the presented result is the average of five runs times the number of subjects.
 
 %does not appear at different images or videos between training, validation and testing set. 

 %Some of the works applied validation techniques separating images or videos disregarding the subject shown in the image. Few works separate subjects(animals) between train, validation and tests sets and allocate all their corresponding images/videos accordingly. 
 
For non-deep methods, Varma et al. \cite{varma2006bias} studied validation techniques  suggesting that Nested Cross-Validation \cite{7376625} has minimally biased performance estimates. Vabalas at al. \cite{vabalas2019machine} also recommended this method to be used with datasets with a sample size of up to 1000, to reduce strongly biased performance. In this method, a portion of the data is split at the beginning and in each cross-validation fold, a model is then developed on the reduced training set from scratch, including feature selection and parameter tuning. This is repeated with splitting a different portion of the data for validation, each time developing a new model for training until all the data is used. 
 Koskimäki \cite{7376625} showed that to obtain more confidence on the results, models should be trained and evaluated applying at least using nested or single 10-fold cross-validation or by using double or simple LOAO cross-validation.  
 
 In a simple leave-one-person-out cross-validation the validation is set randomly. In double (or nested) leave-one-person-out cross-validation, bias is avoided by adding an outer loop into cross validation. Data from one person at a time is chosen as separate testing data while the data from the remaining N-1 subjects is left for basic leave-one-person-out cross-validation. This approach is, however, the most computationally challenging.  Nested 10-fold cross-validation or double leave-one-person-out  methods are recommended to reduce biased performance when feature selection or parameter tuning is performed during cross-validation (Table \ref{tab:crossrec}).

Therefore, in small datasets (number of samples lower than 1000), which have almost no repetition of subjects, 10-fold Cross-Validation is recommended to reduce biased performance whenever neither feature selection nor parameter tuning is performed during cross-validation. Otherwise, Nested 10-fold Cross-Validation is recommended. For relatively small datasets with numerous repeated samples of same animal subject, the leave-one-animal-out cross-validation technique is recommended to reduce biased performance. Otherwise, Double leave-one-animal-out cross-validation is recommended.

\begin{description} \item [Recommendation 2:] 
To reduce biased performance evaluation, for classical machine learning methods, the choice of cross-validation is recommended according to Table \ref{tab:crossrec}, when the dataset is small with repeated samples of the same animal subject.  For deep methods, it is recommended to use a fully held-out test set, which ideally is subject-exclusive. It is furthermore recommended to present results from repeated runs on more than one random seed.

 \end{description}
 
\begin{table}[]
\centering
\caption{Recommendations for best practices in cross-validation for classical machine learning methods.}
\begin{tabular}{lll}
\hline
\multicolumn{1}{|l|}{\multirow{2}{*}{Feature Selection or Parameter Finetuning Used?}} & \multicolumn{1}{l|}{\multirow{2}{*}{\textbf{No}}} & \multicolumn{1}{l|}{\multirow{2}{*}{\textbf{Yes}}} \\
\multicolumn{1}{|l|}{}                                                    & \multicolumn{1}{l|}{}                             & \multicolumn{1}{l|}{}                              \\ \hline
Repeated samples per subject?                                             &                                                   &                                                    \\ \hline
\multicolumn{1}{|l|}{\textbf{No}}                                         & \multicolumn{1}{l|}{10-fold}                      & \multicolumn{1}{l|}{Nested 10-fold}                \\ \hline
\multicolumn{1}{|l|}{\textbf{Yes}}                                        & \multicolumn{1}{l|}{LOAO}                  & \multicolumn{1}{l|}{Double LOAO}                          \\ \hline
\end{tabular}
\label{tab:crossrec}
\end{table}
 
 \subsection{Domain Transfer}
 
 The variety of species, affective states and environment conditions lends itself to exploration of cross-database transfer methodologies \cite{ShanFacialBias2020}, i.e., training a model on an original, source dataset and subsequently use this instance to classify samples from a target dataset, presenting some degree of domain shift. %this trained model samples of a different dataset (target) in a (sometimes slightly) different domain. 

One possible setting for domain transfer is cross-species. Hummel et al. \cite{hummel2020automatic}, studies domain transfer from horse-based models
to donkeys, reporting a loss of accuracy in automatic
pose estimation, landmark detection, and subsequent pain prediction. %\az{Sofia - please check this!!}
A further example of domain transfer is cross-state: to transfer between different types of affective states or types of pain. In the study of Broom\'e et al. \cite{broome2019dynamics}, it was shown that a model trained only on a dataset of horses with acute experimental pain can aid recognition of the subtler displays of orthopedic pain. This is useful because training is shown to be difficult on the subtler type of pain expression. A third example of a transfer scenario is cross-environment, or simply cross-domain. Lu et al. \cite{mahmoud2018estimation} train their model for pain estimation in sheep on a dataset collected on a farm, and then present results transferred to a dataset of sheep collected from the internet.

Learning from cross-database transfer in human facial analysis, one crucial issue is the differences in intrinsic bias of the source and target datasets, related not only to the facial expressions, but also to important factors such as occlusion, illumination, and background, as well as factors related to annotation and balance, which may have significant impact. Li et al.~\cite{ShanFacialBias2020} demonstrate such differences in the human domain and propose methods to minimize these types of biases. For animals, these differences are expected to play an even greater role, given the large domain variety, as discussed in Section \ref{sec:data}.

% Learning from cross-database transfer in human facial analysis, one crucial issue is the differences in intrinsic bias of the source and target datasets, and the differences between these. Li et al.~\cite{ShanFacialBias2020} investigated this problem in human facial analysis, experimenting with cross-database generalization on specific expression categories. The authors showed that facial expression datasets appear to have a strong built-in bias, and that even deep models are insufficient to generalize well across various datasets. 
% The authors further analyzed the dataset bias quantitatively,  to investigate the reasons for this.
% Specifically, experimental results for facial expression recognition demonstrate that the datasets tend to have their own specific characteristics, stemming from the dataset creation and annotation process. This bias can potentially lead to discrepancy in the final predictions. The experiment further revealed that annotators in each dataset tended to have inconsistent perceptions on the expression categories. This means that the assumption that the conditional distribution %(probability of the data given the model parameters) 
% remains unchanged across domains fails to hold in human emotion recognition, and that it is even less likely to hold for animals due to an even greater domain diversity. A method for overcoming such bias, introduced in \cite{ShanFacialBias2020} is based on the idea of balancing the divergence between the source and target domains by introducing 
% additional learnable class-wise weighting parameters. 

\begin{description} \item [Recommendation 3:] Methods to minimize intrinsic dataset bias are recommended for cross-domain transfer studies.  
 \end{description}

%A method to overcome such bias was presented in their paper. For that reason, we strongly recommend taking this intrinsic dataset bias consideration while utilizing transfer methods.

\section{Conclusions and Future Work}
\label{sec:discussion}

Although the field of automated recognition of affective states in animals is only beginning to emerge, the breadth and variability of the approaches covered in our survey makes this
a timely moment for reflection on challenges faced by the community and steps that can be taken to advance the field. 

One crucial issue that needs to be highlighted is the difficulty in comparing the different works. 
 Despite some commonalities in the stage of data analysis (features, models and pipelines), the variety of species, and the ways data are collected and annotated differ tremendously, as discussed in Section \ref{sec:data}. Thus, e.g.,
the 99\% accuracy achieved in \cite{andresen2020towards} for pain recognition in laboratory mice, in a small box with controlled lighting and good coverage of the animal's face, cannot be straightforwardly compared to the estimation
of pain level with accuracy of 67\% achieved in \cite{mahmoud2018estimation} for sheep using footage obtained in the naturalistic (and much less controlled) setting of a farm.

 Drawing inspiration from the huge amount of benchmark datasets existing in the human domain (such as the Cohn-Kanade dataset \cite{lucey2010extended}, the Toronto face database \cite{susskind2008generating}, and more), the development of benchmarking resources for animals -- both species-specific and across-species -- can help systematize the field and promote comparison between approaches. However, this is more challenging than in humans, due to the large variety of species, and environments (laboratory, zoo, home, farm, in the wild, etc.) in which they can be recorded. Another barrier is considerations of ethics and privacy, especially when producing datasets with induced affective states and pain (as explained in Section \ref{sec:data}). Unfortunately, this often makes it difficult to make datasets publicly accessible. Thus, there is a strong need for public datasets in this domain.   %Further, in biological sciences, it is not as common to publish datasets as in computer science, and the data is often collected on the biological side in an interdisciplinary collaboration. These factors taken together often makes it difficult to make these datasets public.

Another issue that should be addressed in future research efforts is explainability, which is particularly important for applied contexts related to clinical decision making and animal welfare management. Consistently with the human affective computing literature, our review reveals the tendency towards `black-box' approaches which use learned features. While using hand-crafted features is indeed less flexible than learning them from data, and may in some cases lead to lower performance, their clear advantage is explainability, having  
more control over the information extracted from a dataset. Learned features, on the other hand, tend to be more opaque, leading to `black-box' reasoning, which does not borrow itself easily for explaining the classification decisions in human-understandable terms (see, e.g., \cite{london2019artificial}). 
{It is possible} to investigate statistical properties of the various dimensions of the feature maps, and study what type of stimuli specific neurons activate maximally for, but th{ese properties are still not} 
%will not be 
conclusive in terms of how features are organized and what they represent. In neural networks, the features are often \textit{entangled}, which complicates the picture more. This is when a given network unit activates for a mix of input signals (e.g., the face of an animal in a certain pose with a certain facial expression and background), but perhaps not for the separate parts of that signal (e.g., the face of the same animal in a different pose, with the same facial expression, but with a different background). There have been \textit{disentanglement} efforts, predominantly unsupervised \cite{DBLP:conf/iclr/HigginsMPBGBML17, KumarDisentangle18, kimmnih18FactorVAE}, within deep learning to reduce these tendencies, since this is, in general, not a desirable feature for a machine learning system. {In terms of animal affect applications, the pain recognition approach of Rashid et al. \cite{rashid2022equine}, includes such self-supervised disentanglement in one part of their modeling pipeline.} However, much development remains before a neural network can stably display control and separation of different factors of variation. This characteristic of neural networks poses difficulty for research that aims to perform exploratory analysis of animal behavior. In particular, it poses high demands on the quality of data and labels, in order to avoid reliance on spurious correlations. 
%Addressing explainability in this domain is particularly important for applied contexts related to clinical decision making and animal welfare management. 

In summary, in the last five years we are witnessing an impressive growth in the number of studies addressing recognition of affective states and pain in non-human animals. Notably, many of these works are carried out by multi-disciplinary teams, demonstrating the intellectual value of collaboration between biologists, veterinary scientists and computer scientists, as well as the increasing importance of computer vision techniques within animal welfare and veterinary science. Efforts invested in pushing the field of animal affective computing forward will not only lead  
to new technologies promoting animal welfare and well-being, but will also hopefully provide new insights for the long-standing philosophical and ethical debates on 
animal sentience. 

\section*{Acknowledgements}
The research was partially supported by the grant from the Ministry of Science and Technology of Israel according to the research project no. 19-57-06007.
The second author was additionally supported by the Data Science Research Center (DSRC), University of Haifa. 
The authors would like to thank Ilan Shimshoni and Shir Amir for their scientific consultations.

\bibliography{sn-bibliography}% common bib file


\end{document}